\documentclass[runningheads]{llncs}
\usepackage[T1]{fontenc}
\usepackage{graphicx}
\usepackage{algpseudocode}
\usepackage{algorithm}
\usepackage{lmodern}       
\usepackage{anyfontsize}   
\usepackage{array}         
\usepackage{hyperref}
\usepackage{subcaption}
\usepackage{caption}
\usepackage{booktabs}

\algnewcommand\algorithmicinput{\textbf{Input:}}
\algnewcommand\Input{\item[\algorithmicinput]}
\algnewcommand\algorithmicoutput{\textbf{Output:}}
\algnewcommand\Output{\item[\algorithmicoutput]}
\begin{document}
\title{SSL-MedSAM2: A Semi-supervised Medical Image Segmentation Framework Powered by Few-shot Learning of SAM2}
%
%
\author{Zhendi Gong\thanks{Corresponding author. \email{zhendi.gong@nottingham.ac.uk} } \and
Xin Chen}

\authorrunning{Z. Gong and X. Chen}
%
\institute{School of Computer Science, University of Nottingham, UK}
\maketitle              
\begin{abstract}
Despite the success of deep learning based models in medical image segmentation, most state-of-the-art (SOTA) methods perform fully-supervised learning, which commonly rely on large scale annotated training datasets. However, medical image annotation is highly time-consuming, hindering its clinical applications. Semi-supervised learning (SSL) has been emerged as an appealing strategy in training with limited annotations, largely reducing the labelling cost. We propose a novel SSL framework SSL-MedSAM2, which contains a training-free few-shot learning branch TFFS-MedSAM2 based on the pretrained large foundation model Segment Anything Model 2 (SAM2) for pseudo label generation, and an iterative fully-supervised learning branch FSL-nnUNet based on nnUNet for pseudo label refinement. The results on MICCAI2025 challenge CARE-LiSeg (Liver Segmentation) demonstrate an outstanding performance of SSL-MedSAM2 among other methods. The average dice scores on the test set in GED4 and T1 MRI are 0.9710 and 0.9648 respectively, and the Hausdorff distances are 20.07 and 21.97 respectively. The code is available via \url{https://github.com/naisops/SSL-MedSAM2/tree/main}.

\end{abstract}
\section{Introduction}
Liver segmentation in MRI plays a crucial role in diagnosing and monitoring a wide range of hepatic diseases, including hepatocellular carcinoma, cirrhosis, and fatty liver disease \cite{bilic2023liver}. Despite the remarkable progress facilitated by deep learning, its full potential in medical image segmentation is frequently affected by the substantial reliance on large-scale, accurately annotated training data \cite{jin2023label}. In clinical practice, obtaining high-quality medical image annotations is an inherently difficult, time-consuming, and expensive effort \cite{gao2025medical}. This contrast has motivated semi-supervised learning (SSL) strategies, which aim to leverage abundant unlabelled data alongside a limited labelled set to train robust segmentation models. 

One common approach is \textbf{pseudo labelling}, where a model’s predictions on unlabelled images are treated as pseudo ground truth to retrain the model or train a student model in an iterative manner. In this branch of approaches, pseudo labels are mostly generated directly from the predictions of the unlabelled data using a trained model followed by some post-processing methods. To increase the reliability of the pseudo labels, recent studies focus on uncertainty-aware methods. For instance, double‑threshold schemes combined classification and segmentation confidences to pick reliable pixels \cite{zeng2023ss}. Ssa-net \cite{wang2022ssa} added a trust module to re‑evaluate model outputs and retain only high‑confidence predictions. Additionally, pseudo labels can also be generated from label propagation by prototype learning \cite{han2022effective} or image registration \cite{xu2019deepatlas}. These approaches aim to transfer the knowledge from the labelled data to the unlabelled ones. As a main constraint of the pseudo labelling strategy, the overall performance relies on the robustness of the initial model for pseudo labelling, and the inaccuracy in pseudo labels can be quickly amplified during the training process.

Another widely used approach is \textbf{consistency regularization}, which encourages a segmentation model to produce invariant predictions under different perturbations of the same input, effectively pushing the decision boundary into low‑density regions of the data manifold. Normally, input-level perturbation like Gaussian noise or mix‑up augmentations to shift input data \cite{li2020transformation,basak2022exceedingly}, or feature-level perturbation to modify intermediate feature maps \cite{li2021dual,ouali2020semi} is introduced. The widely used framework in this branch of methods is mean teacher \cite{tarvainen2017mean}, which used a ``teacher” network whose weights are the exponential moving average (EMA) of the ``student” network weights. Consistency is enforced between the student and teacher predictions on perturbed inputs. Beyond data‑level augmentations, some methods imposed consistency between related tasks, by adding auxiliary task  (e.g. signed distance map regression, contour prediction, reconstruction) to leverage geometric information \cite{chen2022semi,li2020shape}. Main limitation of the consistency regularization methods is the dependency on the strength and type of perturbations. Weak perturbations may yield little benefit, while overly strong ones can confuse the model and degrade performance \cite{jiao2024learning}.

Most existing SSL methods train models from scratch on the small labelled set, which may limit their ability to quickly capture generalizable features from the outset. Recently, the emergence of large foundation models for image segmentation offers a new opportunity to address these challenges due to their impressive zero-shot and few-shot segmentation capabilities \cite{miao2024cross}. In particular, the Segment Anything Model (SAM) \cite{kirillov2023segment} has garnered attention as a promotable segmentation model trained on billions of 2D natural image segmentation tasks, capable of generalizing to a wide range of objects with minimal user input (e.g. points or bounding boxes).

Building on SAM, Segment Anything Model 2 (SAM2) \cite{ravi2024sam} extended its capabilities to video segmentation, by treating 2D images as video frames, and proposing a streaming memory mechanism to transfer knowledge between frames. The spatial continuity between adjacent slices in a 3D medical image parallels the temporal coherence of successive video frames. By treating a 3D volume data as a video, the adaption of SAM2 on medical images domain is more robust than SAM \cite{zhu2024medical}. By fine-tuning SAM2 on large-scale medical datasets, MedSAM2 \cite{ma2025medsam2} had shown superior performance on plenty of medical segmentation tasks.

Deploying such a foundation model in medical imaging could allow rapid learning from very limited labelled data due to its reliable predictions from few-shot segmentation, potentially alleviating the need for large annotated datasets \cite{li2023segment}.  In the branch of \textbf{pseudo labelling} approaches in SSL methods, based on the prompts generated by SSL models, SAM and SAM2 can be applied to select reliable pseudo-labels, showing a superior performance compared with existing SSL algorithms. SAMDSK \cite{zhang2023samdsk} used SAM with empty prompt on unlabelled data and selected the reliable predictions by a CNN trained on the labelled dataset. Li et al. \cite{li2023segment} first applied CNN to generate a rough prediction on an unlabelled image, then converted the prediction into point prompts for SAM to produce reliable pseudo masks. To modify the prompt generation process, CPC-SAM \cite{miao2024cross} proposed a cross prompting consistency method to automatically generate reliable prompts for unlabelled data across two decoder branches. Furthermore, in \textbf{consistency regularization} branch, SemiSAM \cite{zhang2024semisam} used a trainable mean teacher framework to produce prompts for SAM, and the predictions from SAM assist in the mean teacher training. 

All the above SAM-based SSL methods cannot directly transfer the valuable knowledge from the labelled data to unlabelled ones by the impressive zero-shot and few-shot segmentation capabilities of SAM or SAM2 without any training. In our method, we propose SSL-MedSAM2, an iterative \textbf{pseudo labelling} approach. Specifically, we first propose a training-free few-shot learning branch TFFS-MedSAM2 based on MedSAM2 \cite{ma2025medsam2}, which adapted SAM2 in medical imaging, to produce initial pseudo labels for the unlabelled data in an ensemble manner. It is then followed by an iterative fully-supervised training branch FSL-nnUNet to refine the pseudo masks. In TFFS-MedSAM2, we leverage SAM2’s promptable interface to perform segmentation on unlabelled images in a few-shot manner. Our proposed method was evaluated using MICCAI 2025 challenge CARELiSeg dataset \cite{liu2025merit,gao2023reliable,wu2022meru} for both the non-contrast and contrast-enhanced subtasks. The contributions of this work are summarized as follows:
\begin{enumerate}
\item We propose SSL-MedSAM2, a novel semi-supervised medical image segmentation framework. To our knowledge, this is one of the first frameworks to uniquely combine a large foundation segmentation model adapted on medical images (MedSAM2) in a training-free manner with a task-specific segmentation network (nnU-Net) into a SSL pipeline. 

\item We introduce TFFS-MedSAM2, a novel training-free few-shot learning branch in an ensemble predicting manner that harnesses the pre-trained MedSAM2 model for efficient and high-quality pseudo-label generation on unlabelled volumetric data without requiring of any user prompts. 

\item Our framework achieves outstanding performance on the CARE-LiSeg challenge for both GED4 and T1 MRI, where only a few GED4 MRI is annotated, demonstrating its generalizability and robustness across different scans and its efficacy and potential for clinical applications by achieving superior performance while significantly reducing labelling costs.

\end{enumerate}

\section{Method}
\subsection{Preliminaries}
SSL aims to train high-performing models based on the combination of limited labelled data and a large amount of unlabelled data. We formulate the SSL task as following. Given a dataset $\mathcal{D}$, it consists of two subsets $\mathcal{D}^l$ and $\mathcal{D}^u$, i.e. $\mathcal{D} = \{\mathcal{D}^l,\mathcal{D}^u\}$. $M$ fully annotated cases constitute the labelled subset $\mathcal{D}^l = \{x_i^l, y_i^l\}_{i=1}^M$, and $N$ unlabelled cases constitute the unlabelled subset $\mathcal{D}^u = \{x_i^u\}_{i=1}^N$ ($M \ll N$). $x_i^l$ and $x_i^u$ denote the labelled and unlabelled input images, respectively, and $y_i^l$ denotes the corresponding ground truth mask of $x_i^l$. Our SSL-MedSAM2 framework aims to generate reliable pseudo masks $\hat{y}_i^u$ for $\mathcal{D}^u$ to build-up $\mathcal{D}_{p}^u = \{x_i^u, \hat{y}_i^u\}_{i=1}^N$ by TFFS-MedSAM2, then $\{\mathcal{D}^l,\mathcal{D}_p^u\}$ can be used for fully-supervised learning to refine $\hat{y}_i^u$ iteratively by FSL-nnUNet.

\subsection{SSL-MedSAM2 Framework}
SSL-MedSAM2 contains a few-shot learning branch TFFS-MedSAM2 based on the pretrained large foundation model MedSAM2 \cite{ma2025medsam2} and an iterative fully-supervised learning branch FSL-nnUNet based on nnUNet structure \cite{isensee2021nnu}. The overall workflow of SSL-MedSAM2 is shown in Algorithm \ref{alg:sslm}, where $f_\theta$ indicates MedSAM2 and $g_\theta$ indicates nnUNet. The first step (Pseudo Label Initialization) is completed by TFFS-MedSAM2, while the remaining steps are performed in FSL-nnUNet. MedSAM2 is used to generate the initial pseudo masks for the unlabelled data with the assistance of the prompts from the labelled data, and nnUNet is used to iteratively refine the pseudo masks from TFFS-MedSAM2. In model inference, only the refined nnUNet ($g_\theta$) is used. Details of TFFS-MedSAM2 and FSL-nnUNet are explained in the following sections.

\begin{algorithm}[!b]
\caption{Training procedure of SSL-MedSAM2.}
\label{alg:sslm}
\begin{algorithmic}
  \State $\displaystyle \hat{Y}^{u} \gets f_\theta(X^u \mid \mathcal{D}^l=(X^l,Y^l))$ \Comment{Pseudo Label Initialization}
  
  \While{not converge}
    \State $\mathcal{D} \gets \{\mathcal{D}^l,(X^u,\hat{Y}^{u})\}$
    \Comment{Training Set Building} 
    
    \State $g_\theta \gets $ train on $\mathcal{D}$
    \Comment{Model Refinement} \label{ag:WFO}
    
    \State $\hat{Y}^{u} \gets g_\theta(X^u)$
    \Comment{Pseudo Label Updating} \label{ag:comp}
  \EndWhile \\
\Return Refined model $g_\theta$
\end{algorithmic}
\end{algorithm}

\begin{figure}[!b]
\centering
\includegraphics[width=\linewidth]{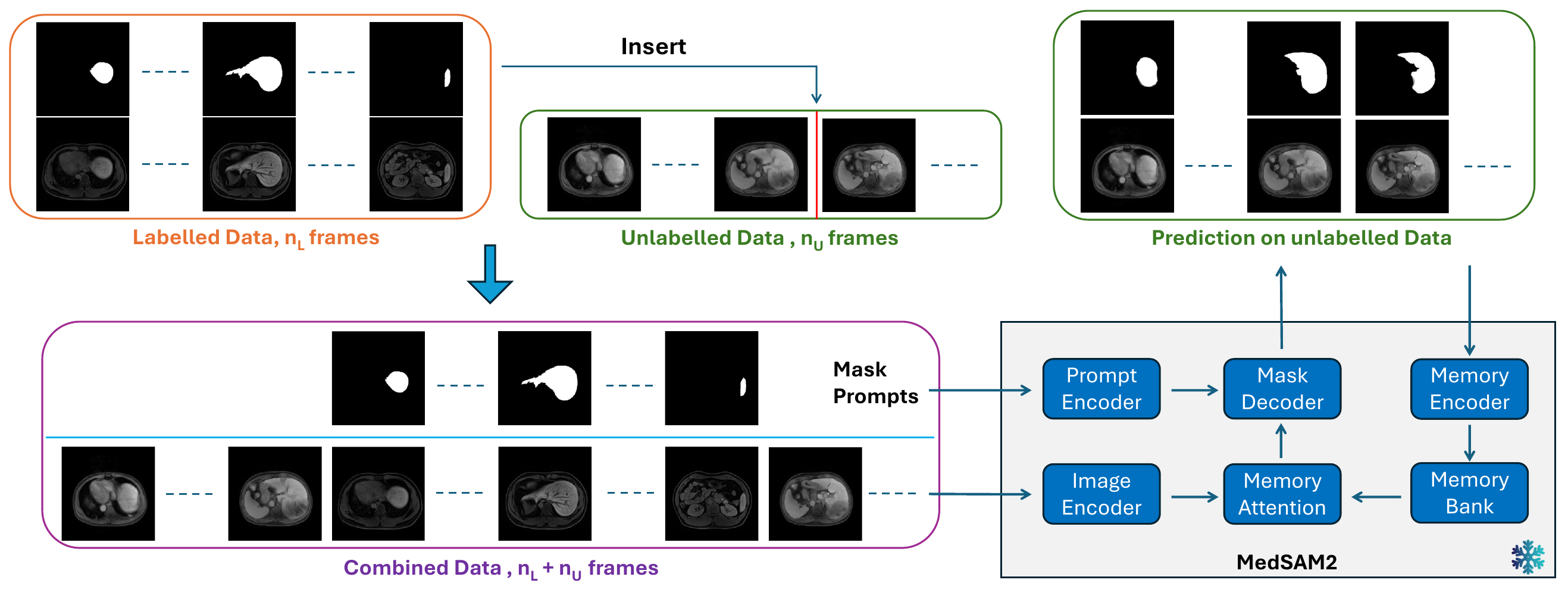}
\caption{The workflow of the composition of labelled data and unlabelled data, and the prediction on the unlabelled data from the composition. The inserting location is denoted by the red line in the unlabelled data. MedSAM2 is frozen in the whole process.}
\label{fig:insert}
\end{figure}
\subsubsection{TFFS-MedSAM2}
SAM2 employs a transformer‑based backbone augmented with a streaming memory module, enabling it to process sequences of frames. While SAM supports point, box, and mask prompts on single images, SAM2 extends these prompts across frame sequences, propagating and refining segmentation masks automatically as the object moves or deforms \cite{ravi2024sam}. The core innovation of SAM2 is a lightweight memory bank that stores embeddings from previous frames, enabling the model to reference past context when segmenting the current frame. SAM2’s streaming memory naturally propagates segmentations across 3D medical volume slices, reducing the need for per‑slice prompts. By leveraging the temporal coherence, SAM2 can produce high‑quality masks for the entire volumes from a minimal input. Thus, we combine the labelled and unlabelled slices together and treat them as a whole video sequence, and use the limited labelled frames as guidance (mask prompts) for SAM2 to predict on the unlabelled frames. Specifically, we use MedSAM2 \cite{ma2025medsam2} that was fine-tuned on the large-scale medical image segmentation tasks instead of the original SAM2 to fully utilize the domain knowledge in the pretrained large foundation model.

As shown in Fig. \ref{fig:insert}, by treating the input 3D volume data as video frames, a whole labelled 3D volume is inserted to an unlabelled volume at random slice locations. The combined frames are then input to the pretrained and frozen MedSAM2 to generate pseudo masks for the unlabelled frames, with the mask prompt provided by the labelled frames (no prompt on the unlabelled frames). With the memory attention that stores the knowledge of the prompt from the labelled frames and the predictions from the adjacent frames, MedSAM2 can detect similar image features of the unlabelled object to the predicted frames and generate predictions through the mask decoder. 

The process of ensemble pseudo labelling of each unlabelled volume by TFFS-MedSAM2 is shown in Algorithm \ref{alg:fs}, where $R$ denotes the number of random inserting locations in each unlabelled data, $x_i^c$ denotes the labelled-unlabelled combined data (purple box in Fig. \ref{fig:insert}), and $p_i^u$ denotes the probability map of $x_i^u$, generated by $f_\theta$ (MedSAM2). In TFFS-MedSAM2, each of the $M$ labelled data is inserted to each unlabelled data at $R$ random locations. Thus, $R \times M$ probability maps are generated for each unlabelled data, the average of the probability maps is used to generate the pseudo mask, improving robustness through prediction ensembling. The whole process is training-free, and the pseudo-labelling of each unlabelled data is independent, requiring no user prompts and no assistance from other unlabelled data, yielding a robust and effective few-shot learning capability. 

\begin{algorithm}[!t]
\caption{Pseudo‑Labelling of unlabelled data \(x_j^u\) by TFFS‑MedSAM2.}
\label{alg:fs}
\begin{algorithmic}[1]
  \Input Unlabelled data \(x_j^u\) and labelled dataset \(\mathcal{D}^l = \{(x_i^l, y_i^l)\}_{i=1}^M\)
  \Output Pseudo‑label \(\hat{y}_j^u\) for \(x_j^u\)

  \State $\hat{p}_j^u \gets 0$ \Comment{Zero Initialization}
  \For{$i \gets 1$ to $M$}
    \State Randomly generate $R$ inserting locations on the unlabelled frames
    \For{$r \gets 1$ to $R$}
        \State $\{x_i^{c},y_i^l\} \gets$ Insert $\{x_i^l,y_i^l\}$ to \(x_j^u\) between the frame index $F^{(r)}$ and $F^{(r)}+1$
        \State $\hat{p}_j^u \gets \hat{p}_j^u + Sigmoid(f_\theta(x_j^u \mid\{x_i^{c},y_i^l\}))$ 
    \EndFor
    \EndFor
    \State $\hat{p}_j^u \gets \hat{p}_j^u /(R \times M)$            \Comment{Get Average}
    \State $\hat{y}_j^u \gets threshould(\hat{p}_j^u)$        \Comment{Get Binary Mask}
  
\end{algorithmic}
\end{algorithm}

\subsubsection{FSL-nnUNet}
After generating the pseudo masks for the unlabelled dataset using TFFS-MedSAM2, the second branch of our framework is an iterative fully-supervised learning process FSL-nnUNet using nnU-Net on $\mathcal{D} = \{\mathcal{D}^l,(X^u,\hat{Y}^{u})\}$ to iteratively refine the pseudo masks $\hat{Y}^{u}$. Specifically, to avoid over-estimation when performing model inference on the unlabelled data for updating pseudo labels, the unlabelled dataset is split into 5 folds, and we perform 5-folds cross-validation using nnUNet. In each fold of training, the validation set is from one fold of the unlabelled dataset, and the remaining part of the unlabelled dataset along with the labelled dataset form the training set. After training, 5 folds predictions on the different validation sets constitute the updated pseudo masks for the whole unlabelled dataset.

Initially, the nnU-Net model is trained on the small set of labelled images together with the MedSAM2-generated pseudo-labels for the unlabelled images. Although the pseudo-labels may be imperfect, the foundation model provides a strong starting point that guides the segmentation network. The newly trained nnU-Net is used to re-predict segmentation masks for the unlabelled data, which results in more refined masks than the original MedSAM2 output due to the network’s adaptation to the specific task. These refined predictions are treated as updated pseudo-labels, and we retrain the nnU-Net using the improved labels. This iterative self-training cycle can be repeated multiple times, continually improving the quality of the pseudo labels and the performance of the segmentation model. All of the final refined 5-folds models are used for model inference on the unseen testing data by taking the average of 5 different predictions. FSL-nnUNet is a flexible framework, where nnUNet can be easily replaced by another fully-supervised medical segmentation method.

\section{Experiments and Results}
\subsection{Dataset and Experimental Settings}
CARE-LiSeg \cite{liu2025merit,gao2023reliable,wu2022meru} dataset consists of T2-weighted imaging (T2WI), diffusion-weighted imaging (DWI), and Gadolinium ethoxybenzyl diethylenetriamine pentaacetic acid (Gd-EOB-DTPA)-enhanced dynamic MRIs. The Gd-EOB-DTPA-enhanced dynamic MRIs cover the non-contrast phase (T1), arterial phase, venous phase, delayed phase, and hepatobiliary phase (GED4) respectively. In the whole dataset, only limited ground truth of GED4 MRI is available. The \textbf{Contrast-Enhanced} subtask focus on the segmentation of GED4, while the \textbf{Non-Contrast} subtask focus on T1, T2WI, and DWI. The whole dataset contains 610 patients diagnosed with liver fibrosis, all of whom underwent multi-phase MRI scans using three different scanners, which are Philips Ingenia3.0T, Siemens Skyra 3.0T, and Siemens Aera 1.5T (Vendor A, B1, and B2) respectively, resulting in a multi-phase and multi-center dataset. Different from the training and validation sets, the testing set also includes cases from an unseen vendor (denoted by Vendor C), evaluating the out-of-distribution (OOD) performance, additional to the in-distribution (ID) performance. The details of the dataset is shown in Table. \ref{tab:data}. 
\begin{table}[!t]
\caption{The data distribution of the CARE-LiSeg dataset. The value in brackets indicates the amount of annotated data in that subset. }
\label{tab:data}
\centering
\tiny
\resizebox{0.75\columnwidth}{!}{%
\begin{tabular}{l|cccc}
\hline
           & Vendor A & Vendor B1 & Vendor B2 & Vendor C \\ \hline
Training   & 130 (10)      & 170 (10)      & 60 (10)       & 0        \\
Validation & 20       & 20        & 20        & 0        \\
Testing    & 40       & 40        & 40        & 70       \\ \hline
\end{tabular}%
}
\end{table}

In TFFS-MedSAM2, all input frames to the pretained MedSAM2 were resized to $512\times512$ due to the default setting. After the model inference of MedSAM2, the predictions were resized back to the original image size for pseudo label generation. In FSL-nnUNet, the training settings were automatically defined by nnUNet. Particularly, all nnUNet models were trained for 1000 epochs, and we ran FSL-nnUNet 3 times for the \textbf{Contrast-Enhanced} subtask and twice for the \textbf{Non-Contrast} subtask. Specifically, when generating the pseudo masks for the data in non-contrast sequences by TFFS-MedSAM2, we still combined them with the labelled GED4 data (although they are in different MRI sequences). All experiments were deployed on a GPU instance with about 10 GB memory from a Nvidia A100 GPU. During evaluation, the mean value of dice coefficient (DC) and Hausdorff distance (HD) were reported.

\subsection{Results and Discussion}
\begin{table}[!t]
\caption{Evaluation results of the validation set in terms of GED4 and T1 reported on leaderboard. Top 5 teams are shown.}
\label{tab:ssl-val}
\centering
\begin{subtable}{0.48\linewidth}
\centering
\begin{tabular}{lll}
\hline
Team  & Dice & HD \\ \hline
TanglangSled  &  97.30    & 31.01   \\
MHIL &  97.05    &  19.55  \\
Ours &   97.04   &  21.28  \\
Monster &   97.04   &  19.28  \\
InteliLiver & 97.00     & 21.34   \\
Ours-TFFS & 95.92     & 24.71   \\ \hline
\end{tabular}
\caption{GED4.}
\end{subtable}
\hfill
\begin{subtable}{0.48\linewidth}
\centering
\begin{tabular}{lll}
\hline
Team  & Dice & HD \\ \hline
Ours  &  95.74    &  41.14  \\
Ours-TFFS &  95.46    &  40.65  \\
LidaYang &  94.88    &  49.10  \\
MHIL &  94.65    &  62.84  \\
Sigma &   94.19   &  59.32  \\
CitySJTU &  93.81    &  44.98  \\
 \hline
\end{tabular}
\caption{T1.}
\end{subtable}
\end{table}

For the Non-Contrast subtask of the CARE 2025 challenge, results were submitted only for the T1 sequence, as the initial pseudo masks generated by TFFS-MedSAM2 for T2WI and DWI sequences were of insufficient quality for direct use in FSL-nnUNet training. This limitation likely arises from the larger appearance discrepancy between GED4 and T2WI/DWI compared to GED4 and T1, given that T1 and GED4 represent two distinct phases of the same Gd-EOB-DTPA-enhanced acquisition, thus sharing more consistent intensity and structural patterns.

The average Dice coefficient and HD on the validation datasets, as reported on the official leaderboard, are summarized in Table~\ref{tab:ssl-val}. For comparison, we also report the performance of the initial pseudo masks generated directly by TFFS-MedSAM2 (denoted as Ours–TFFS), prior to any refinement by FSL-nnUNet. The corresponding results on the testing datasets are presented in Table~\ref{tab:ssl-test}. Compared to the validation results, the testing phase also reported the In-Distribution (ID) and Out-Of-Distribution (OOD) performance separately to evaluate the robustness of the methods on the unseen vendor data. However, due to restrictions on the number of submissions permitted during the testing phase, the results of Ours–TFFS were not evaluated on the test server. 

As shown in Table~\ref{tab:ssl-val}, our method ranks third in both Dice and HD for the GED4 validation data, while achieving the best overall performance for the T1 validation data. Notably, the Ours–TFFS results indicate that even without the iterative refinement stage, TFFS-MedSAM2 alone can generate high-quality pseudo masks through its training-free few-shot segmentation mechanism — even outperforming other participants on the T1 data. This demonstrates the strong initialization capability of the few-shot branch and its effectiveness in transferring structural priors from limited annotated volumes. Furthermore, the inclusion of FSL-nnUNet refinement substantially improves segmentation accuracy for the GED4 sequence, whereas the gain is less pronounced for T1.

\begin{table}[htbp]
\caption{Evaluation results of the test set in terms of GED4 and T1 reported on leaderboard. ``In-Distribution'' contains data from vendor A, B1, B2. ``Out-Of-Distribution'' contains data from vendor C. Top 5 teams are shown.}
\label{tab:ssl-test}
\centering
\begin{subtable}[t]{0.48\textwidth}
\centering
\begin{tabular}{lll}
\hline
Team  & Dice & HD \\ \hline
Monster  &  97.03    & 21.40   \\
Ours &  96.84    &  22.97  \\
xcj &   96.47   &  25.03  \\
BI &   96.44   &  22.63  \\
TeamSpace & 96.41     & 24.32   \\ \hline
\end{tabular}
\caption{In-Distribution GED4.}
\end{subtable}
\hfill
\begin{subtable}[t]{0.48\textwidth}
\centering
\begin{tabular}{lll}
\hline
Team  & Dice & HD \\ \hline
SuperIdols  &  97.70    &  14.65  \\
Monster &  97.65    &  15.17  \\
Ours &  97.54    &  15.11  \\
BI &   97.54   &  14.37  \\
TeamSpace &  97.53    &  15.72  \\ \hline
\end{tabular}
\caption{Out-Of-Distribution GED4.}
\end{subtable}

\vspace{1em}

\begin{subtable}[t]{0.48\textwidth}
\centering
\begin{tabular}{lll}
\hline
Team  & Dice & HD \\ \hline
Ours  &   96.08   &  19.87  \\
BIGS2 &  94.34    &  38.06  \\
Sigma &   93.18   &  58.99  \\
CitySJTU &  92.62    &  27.80  \\
BioDreamer &  91.96    &  31.83  \\ \hline
\end{tabular}
\caption{In-Distribution T1.}
\end{subtable}
\hfill
\begin{subtable}[t]{0.48\textwidth}
\centering
\begin{tabular}{lll}
\hline
Team  & Dice & HD \\ \hline
Ours  &  97.16    &  25.58  \\
BIGS2 &  95.48    &  22.11  \\
Sigma &   95.03   & 29.04   \\
CitySJTU &   94.44   & 25.54   \\
BioDreamer &   94.18   &  28.24  \\ \hline
\end{tabular}
\caption{Out-Of-Distribution T1.}
\end{subtable}
\end{table}

On the testing datasets, our method consistently ranks among the top three approaches across all leaderboard categories. As shown in Table~\ref{tab:ssl-test} (a) and (b), SSL-MedSAM2 achieves top-three performance on the GED4 data in both ID and OOD settings with respect to both Dice and HD metrics. More notably, Table~\ref{tab:ssl-test} (c) and (d) show that our method obtains the highest Dice score and lowest HD among all participants for most cases on the T1 testing data, except for the OOD HD metric, where it ranks third.

Overall, these results confirm that SSL-MedSAM2 delivers SOTA performance across both validation and testing datasets. In particular, its superior accuracy on the T1 sequence demonstrates excellent knowledge transfer between MRI sequences captured in different acquisition phases of the same contrast agent (Gd-EOB-DTPA-enhanced). Moreover, the model exhibits strong OOD generalization, often performing better on unseen vendor data than on the training-domain data. This robustness to domain shifts highlights the effectiveness of combining a foundation segmentation model (MedSAM2) with iterative semi-supervised refinement, making SSL-MedSAM2 a promising solution for practical clinical applications where data heterogeneity is inevitable.

\section{Conclusion}
In summary, we propose a novel semi-supervised learning medical image segmentation framework SSL-MedSAM2 containing a training-free few-shot learning branch TFFS-MedSAM2 and an iterative fully-supervised learning branch FSL-nnUNet. The evaluation results demonstrate the generalizability and robustness of our method even when labelled dataset and the unlabelled dataset are in different phases (e.g, GED4 and T1). However, when the data distribution of the support labelled data and query unlabelled data differs too much (e.g., GED4 and T2WI), TFFS-MedSAM2 tends to generate low-quality pseudo masks. Thus, in the future, we will focus on cross-domain reliable pseudo mask generation and selection in TFFS-MedSAM2 to further enhance the generalizability and robustness of our method. 
%
%
%
\bibliographystyle{splncs04}
\bibliography{mybibliography}

\end{document}